\documentclass[letterpaper, 10 pt, conference]{ieeeconf}  

\IEEEoverridecommandlockouts                              
\usepackage{graphics} 
\usepackage{epsfig} 
\usepackage{mathptmx} 
\usepackage{times} 
\usepackage{amsmath} 
\usepackage{amssymb}  
\usepackage{booktabs} 
\usepackage[table]{xcolor} 
\usepackage{tabularx}

\title{\LARGE \bf
Risk Informed Reinforcement Learning with Gibbs Priors for Topology Control in Power Grids
}

\author{Pantelis Dogoulis and Maxime Cordy%
\thanks{Both authors are with SerVal, SnT,
        University of Luxembourg, Luxembourg City, Luxembourg
{\tt\small \{panteleimon.dogoulis, maxime.cordy\}@uni.lu}}%
}

\begin{document}

\maketitle
\thispagestyle{empty}
\pagestyle{empty}

\begin{abstract}
Topology control for power grid operation is a challenging sequential decision making problem because the action space grows combinatorially with the size of the grid and action evaluation through simulation is computationally expensive. We propose a risk-informed Reinforcement Learning framework that combines semi-Markov control with a Gibbs prior, that encodes the system's physics, over the action space. The decision is only taken when the grid enters a hazardous regime, while a graph neural network surrogate predicts the post action overload risk of feasible topology actions. These predictions are used to construct a risk-informed Gibbs prior that both selects a small state-dependent candidate set and reweights policy logits before action selection. In this way, our method reduces exploration difficulty and online simulation cost while preserving the flexibility of a learned policy. We evaluate the approach in three realistic benchmark environments of increasing difficulty. Across all settings, the proposed method achieves a strong balance between control quality and computational efficiency: it matches oracle-level performance while being approximately $6\times$ faster on the first benchmark, reaches $94.6\%$ of oracle reward with roughly $200\times$ lower decision time on the second one, and on the most challenging benchmark improves over a PPO baseline by up to $255\%$ in reward and $284\%$ in survived steps while remaining about $2.5\times$ faster than a strong specialized engineering baseline. These results show that our method provides an effective mechanism for topology control in power grids.
\end{abstract}
\section{INTRODUCTION}
Modern power grids operate close to their thermal and security limits due to growing variability in generation and demand, increased transfers, and tighter operational margins. In this regime, a large fraction of operational risk is associated with constraint-based dynamics: local overloads can trigger protective line trips, induce non-local flow redistribution, and potentially initiate cascading failures. Operators mitigate such events by combining corrective controls (e.g. topology reconfiguration) with preventive and emergency procedures, under  feasibility constraints (cooldowns, maintenance) and under uncertainty in future injections and contingencies. Designing decision-making algorithms for this setting is difficult because the system is high-dimensional, the dynamics are partially driven by exogenous and stochastic factors, and most of the discrete actions that an operator can perform are inherently combinatorial in size.

Reinforcement Learning (RL) has recently emerged as a promising framework for the power grid operation problem, where the agent interacts with a simulator that encodes power-flow physics and operational constraints (\cite{marot2021learning,kelly2020reinforcement,sar2025optimizing,subramanian2021exploring}). However, this framework exposes a fundamental algorithmic problem: while the environment provides rich state information (flows, voltages, topology, constraints), the action space induced by topology configuration changes grows combinatorially with the network size and substation complexity (see Proposition 1 in \cite{chauhan2023powrl}). As a consequence, naive RL approaches face severe exploration and sample-efficiency challenges, and many practical solutions therefore reduce or restructure the control space before or during learning. Examples include precomputing reduced action sets through extensive simulations \cite{chauhan2023powrl}, integrating domain knowledge and explicit action-space reduction into training \cite{matavalam2022curriculum}, and learning topology-action policies from offline imitation targets derived from simulator outcomes (\cite{hassouna2025learning,de2025generalizable}). Although these approaches improve tractability, they are expensive to curate, environment-specific, and can bias the learned policy toward a fixed action space or fixed supervision signal that may not remain optimal across operating points and contingencies.

Furthermore, many approaches rely on online simulation or search to evaluate candidate actions before committing to them (\cite{marot2021learning,zhou2021action,dorfer2022power}). This can be interpreted as a form of receding-horizon search in a discrete action space, but it requires repeated power flow evaluations and therefore becomes expensive when the candidate set is large or when decisions must be made frequently. In practice, this creates a trade-off between safety and robustness, which improve when more candidate actions are simulated, and runtime, which deteriorates as the number of simulations increases. It is important to note that the computational burden is concentrated in stressed operating regimes: when the system is far from limits, the optimal control is often to do nothing, whereas near overloads the decision quality is crucial. Overall, these observations motivate a decision architecture that: \textit{(a)} intervenes only when necessary and \textit{(b)} restricts online evaluation to a dedicated, state-dependent action subset.

In this work, we propose a risk-informed RL method for topology control that addresses the combinatorial action-space and the online simulation problems simultaneously. The method is based on two principles. First, following prior literature, we formulate the problem as a semi-Markov decision process \cite{yoon2021winning}, where the agent intervenes only at times when the grid enters a hazardous regime. Second, at each hazardous decision time we construct a \emph{time-dependent} candidate set by combining \textit{(i)} a graph-based policy with \textit{(ii)} a risk-informed prior that ranks actions by a predicted post-action overload risk. The candidate set is obtained via a $\mathrm{TopK}$ selection under the prior and feasibility constraints (i.e. the best $\mathrm{K}$ actions), and the final action distribution is formed by reweighting policy logits with prior log scores. The prior itself is learned as a one-step risk surrogate trained on simulator rollouts: it predicts the next-step maximum loading ratio after applying an action, using graph neural networks (GNNs) that encode the current dynamics of the system. 

The main contributions of this paper are:
\begin{itemize}
\item We introduce a risk-informed  methodology that selects a small, state-dependent set of feasible actions that improves scalability in large discrete action spaces.
\item We propose an algorithm-agnostic policy rule, that reweights any agent’s action scores with a risk-informed prior that introduces information from the environment.
\item We demonstrate the effectiveness of the proposed approach in real-world power grid operation settings and show that it generalizes across different power grid environments.
\end{itemize}

\section{RELATED WORK}
\label{sec:related_work}

Topology control for power grid operations sits at the intersection of high-dimensional decision making and hard physical safety constraints, which makes it a challenging setting for RL. A major benchmark driving progress in this area is the Learning to Run a Power Network (L2RPN) competition, which provides realistic simulated power grid environments for evaluating sequential decision-making methods under operational constraints (\cite{marot2021learning,sar2025optimizing,kelly2020reinforcement}), and which over successive editions, has introduced a range of realistic grid environments and operational challenges. Complementary comparative studies have also contrasted advanced rule-based and RL-based topology control agents on the same class of environments \cite{lehna2023managing}. As highlighted in the literature, successful approaches typically depart from pure end-to-end RL and instead introduce structure that makes learning and deployment tractable by: \emph{(i)} reducing or restructuring the action space offline, \emph{(ii)} using online simulation or search at decision time, and \emph{(iii)} introducing temporal or hierarchical abstractions over topology changes. Below, we review these directions accordingly.

\paragraph{Offline action-space reduction and offline supervision}
A common approach to the combinatorial action space is to reduce or restructure it before deployment. PowRL identifies important substations and constructs a reduced action set through extensive simulations \cite{chauhan2023powrl}. Curriculum-based RL likewise injects domain knowledge and explicit action-space reduction into training \cite{matavalam2022curriculum}. Closely related work replaces online search with offline supervision: soft-label imitation learning derives training targets from simulated action outcomes \cite{hassouna2025learning}, and recent GNN-based studies analyze topology-control policies under imitation learning with an emphasis on representation and generalization \cite{de2025generalizable}. These approaches improve tractability, but they inherit bias from the offline action space or supervision signal and may miss useful interventions under distribution shift.

\paragraph{Online simulation and planning}
Another line of work keeps a learned proposal mechanism but validates or ranks actions with the simulator at decision time. Action-Set-Based Policy Optimization (SAS) makes this idea explicit: a policy proposes an action set and the simulator filters it to enforce constraints, which yields a non-differentiable selection stage handled with black-box optimization \cite{zhou2021action}. Planning-based approaches push this further by performing search over topology interventions, as in AlphaZero-style topology optimization \cite{dorfer2022power}. Such methods can be highly effective, but their runtime scales with the number of candidate actions and simulator evaluations.

\paragraph{Hierarchical and temporally abstract control}
A third direction changes the decision abstraction itself. SMAAC adopts a semi-Markov view, intervenes only in hazardous regimes, and lets a high-level policy select a goal topology that is executed via lower-level primitive actions \cite{yoon2021winning}. HUGO similarly elevates the decision space from primitive local switches to robust target topologies \cite{lehna2024hugo}. Other approaches factorize control across levels or agents: hierarchical RL decomposes the problem into whether to intervene, where to act, and which local configuration to apply \cite{manczak2023hierarchical}, while hierarchical MARL distributes control across multiple agents to improve scalability in large action spaces \cite{van2023multi}. These abstractions reduce the effective branching factor, but they also introduce bias through the chosen factorization and may still rely on rules or handcrafted coordination.
\section{METHOD}
\label{sec:method}

\subsection{Problem setting}
\label{sec:problem}
We consider power-grid operation as a sequential decision problem in the Grid2Op environment \cite{grid2op}. At each simulator step $t$ (here we use it as  \emph{micro-step}), the environment returns an observation $o_t$ containing the current grid topology, electrical quantities (e.g., line loadings and voltages), and operational constraints (e.g., cooldown timers). The controller selects an action $a_t$ from a discrete action library $\mathcal{A}$, and the simulator returns $(o_{t+1}, r_t, d_t)$, where $r_t$ is a scalar reward and $d_t \in \{0,1\}$ indicates episode termination (e.g. a normal episode end or a blackout).

\subsection{Hazard-gated semi-MDP formulation}
\label{sec:hazard}
In power-grid operation, interventions are only critical when the system approaches thermal or security limits. We therefore adopt a hazard-gated control scheme, following \cite{yoon2021winning}, in which the learned agent acts only at \emph{hazard decision times}. This induces a semi-Markov decision process (semi-MDP). Let $\rho_\ell(o_t)$ denote the loading ratio of line or transformer $\ell$. We define the instantaneous risk proxy as:
\begin{equation}
\mathrm{Risk}(o_t) := \max_{\ell \in \mathcal{L}} \rho_\ell(o_t),
\label{eq:risk}
\end{equation}
and the hazard indicator:
\begin{equation}
h_t := \mathbb{I}\!\left[\mathrm{Risk}(o_t) \ge \delta_h\right],
\label{eq:hazard}
\end{equation}
where $\delta_h > 0$ is a threshold parameter.

When $h_t=0$, our control mechanism is not invoked and the environment evolves under the no-op (no operation) action, while when $h_t=1$, selects an intervention action.
Thus, we have:
\begin{equation}
a_t =
\begin{cases}
a_t^{\mathrm{RL}}, & h_t = 1,\\
a_t^{\mathrm{def}}, & h_t = 0.
\end{cases}
\end{equation}
Let $\{t_m\}_{m\ge 0}$ denote the random sequence of hazard decision times. Given a hazard time $t_m$, we define the next macro-decision boundary by:
$
t_{m+1} := \inf\{t>t_m:\ h_t = 1 \text{ or } d_t = 1\}.
$
Between $t_m$ and $t_{m+1}$, the environment evolves for
$k_m := t_{m+1}-t_m$ micro-steps. We define the corresponding macro-reward as:
\begin{equation}
R_m := \sum_{j=0}^{k_m-1} \gamma^j r_{t_m+j},
\label{eq:macro_reward}
\end{equation}
and the macro-discount as:
\begin{equation}
\Gamma_m := \gamma^{k_m}.
\end{equation}
This converts the original micro-step process into a semi-MDP indexed by $m$, where learning and planning are performed only at hazard times (\textit{macro-steps}).

\subsection{Action library and feasibility mask}
\label{sec:actions}
We construct a finite action library: $\mathcal{A} = \{a^{(0)}, a^{(1)}, \dots, a^{(|\mathcal{A}|-1)}\}$,
where $a^{(0)}$ denotes the no-op action. The remaining actions are unitary grid operations such as substation topology changes and line-status toggles. At each state, we define a state-dependent feasibility set $\mathcal{F}(o_t) \subseteq \mathcal{A}$,
implemented by a boolean mask. 
All subsequent candidate selection and policy evaluation steps are restricted to $\mathcal{F}(o_t)$.

\subsection{Graph representation of grid state}
\label{sec:graph}
A power grid is naturally represented as a graph. We therefore map each observation $o_t$ to a graph object $G_t=(V,E_t)$.

\paragraph{Nodes (busbar-level)}
We use a busbar-level representation. Each substation has $B=2$ busbars, so the node set is fixed:
\begin{equation}
V = \{(s,b)\,:\, s \in \{1,\dots,n_{\mathrm{sub}}\},\; b \in \{1,\dots,B\}\}.
\end{equation}

\paragraph{Edges}
For each \emph{connected} line or transformer $\ell$ linking busbar nodes $u$ and $v$, we add two directed edges $(u\!\rightarrow\!v)$ and $(v\!\rightarrow\!u)$. Disconnected lines contribute no edges, so topology changes are directly reflected in the time-varying edge set $E_t$.

\paragraph{Features}
The node features $x_u \in \mathbb{R}^{4}$,  include net active and reactive injections $(P_{\mathrm{net}},Q_{\mathrm{net}})$, a busbar voltage estimate, and the substation cooldown associated with that busbar. The edge features $e_{uv} \in \mathbb{R}^{5}$, include the loading ratio $\rho_\ell$ directed active and reactive flows $(P_{uv},Q_{uv})$, the sending-end voltage for the directed edge, and the line cooldown indicator.

\subsection{GNN surrogate}
\label{sec:gnn}
We employ an adapted version of KCLNet introduced in our previous work \cite{dogoulis2025kclnet}. The updated encoder consists of an edge-MLP message-passing component, edge-aware attention refinement and a learnable skip connection. 
\paragraph{Edge-MLP message passing}
We compute messages as:
\begin{equation}
m_{j\to i} = \mathrm{MLP}_{\mathrm{msg}}\!\big([x_j, x_i, e_{ji}]\big),
\end{equation}
and aggregate them by summation:
\begin{equation}
\tilde{h}_i = \sum_{j \in \mathcal{N}(i)} m_{j\to i},
\label{eq:sum_agg}
\end{equation}
where $\mathcal{N}(i)$ is the set of incoming neighbors of $i$.

\paragraph{Edge-aware attention refinement}
We refine $\tilde{h}_i$ using multi-head attention with logits depending on $(\tilde{h}_j,\tilde{h}_i,e_{ji})$:
\begin{align}
\alpha_{j\to i}^{(h)} &= \mathrm{SoftMax}_{j \in \mathcal{N}(i)}
\Big(\mathrm{LeakyReLU}\big(w_h^\top [\tilde{h}_j,\tilde{h}_i,e_{ji}]\big)\Big), \\
\hat{h}_i^{(h)} &= \sum_{j \in \mathcal{N}(i)} \alpha_{j\to i}^{(h)} W_h \tilde{h}_j, \\
h_i &= \frac{1}{H}\sum_{h=1}^{H} \hat{h}_i^{(h)}.
\end{align}

\paragraph{Skip connection and pooling}
We use a learnable skip connection of the form:
\begin{equation}
h_i \leftarrow h_i + W_{\mathrm{skip}} x_i.
\end{equation}
We stack two such blocks. A graph-level embedding is then obtained by mean pooling:
\begin{equation}
z(G_t) = \frac{1}{|V|}\sum_{i\in V} h_i.
\label{eq:pool}
\end{equation}

\subsection{Risk prior}
\label{sec:prior}
We introduce a risk-informed prior over actions that favors interventions predicted to reduce overload risk. This consists of two parts: \textit{(i)} a one-step action-conditional surrogate for post-action risk, and \textit{(ii)} a risk score derived from that surrogate.

\paragraph{One-step risk surrogate}
We learn a parametric surrogate $\widehat{\mathrm{Risk}}_\phi(G_t,a)$ to predict the \emph{next-step} risk resulting from action $a$:
\begin{equation}
\widehat{\mathrm{Risk}}_\phi(G_t,a)\approx \mathrm{Risk}(o_{t+1}^{(a)}),
\end{equation}
where $o_{t+1}^{(a)}$ denotes the next observation obtained after applying action $a$ at time $t$. We implement $\widehat{\mathrm{Risk}}_\phi$ as:
\begin{equation}
\widehat{\mathrm{Risk}}_\phi(G_t,a)
=
\mathrm{ReLU}\!\Big(\mathrm{MLP}_{\mathrm{risk}}([z_\phi(G_t), e(a)])\Big),
\label{eq:risk_hat}
\end{equation}
where $z_\phi(G_t)$ is the graph embedding produced by the GNN encoder and $e(a)$ is a learnable embedding of the action.

\paragraph{Risk score}
We define the risk score as:
\begin{equation}
s_\phi(a\mid G_t)
:=
-\frac{1}{\tau}\widehat{\mathrm{Risk}}_\phi(G_t,a),
\label{eq:prior}
\end{equation}
with temperature $\tau>0$, which controls the sharpness of the risk prior, with smaller values concentrating probability mass more strongly on actions with lower predicted risk. Then, the corresponding normalized Gibbs prior is:
\begin{equation}
p_{\mathrm{phys}}(a\mid G_t) \propto \exp\!\big(s_\phi(a\mid G_t)\big).
\end{equation}
For a fixed $G_t$, the Gibbs form can be interpreted as the maximum entropy distribution over actions under a constraint on the expected predicted risk $\mathbb{E}_{a\sim p(\cdot\mid G_t)}[\widehat{\mathrm{Risk}}_\phi(G_t,a)]$ \cite{jaynes1957information}.

\subsection{State-dependent action set}
\label{sec:topk}
At hazard times $t_m$, we form a state-dependent candidate action set by selecting the $K$ best feasible actions ($\mathrm{TopK}$) under the risk score, which we define as:
\begin{equation}
\mathcal{A}_{t_m}^K
=
\mathrm{TopK}_{a \in \mathcal{F}(o_{t_m})}\; s_\phi(a\mid G_{t_m}).
\label{eq:topk}
\end{equation}
Equivalently, this selects the $K$ feasible actions with the smallest predicted risks $\widehat{\mathrm{Risk}}_\phi(G_{t_m},a)$. 

This reduces the effective action space from $|\mathcal{A}|$ to $K \ll |\mathcal{A}|$ in a state-dependent manner, while explicitly biasing candidates towards physically plausible and risk-reducing interventions.

\subsection{Risk-reweighted policy on the candidate set}
\label{sec:posterior}
We learn a policy-value network using a second graph encoder of the form in \S\ref{sec:gnn}. Given $G_t$, the network outputs policy logits $\ell_\theta(a\mid G_t)$ over the full action set and a scalar value estimate $V_\theta(G_t)$. At hazard times, we restrict the policy to $\mathcal{A}_t^K$ and combine the candidate logits with the risk score:
\begin{equation}
\tilde{\ell}(a\mid G_t)
=
\ell_\theta(a\mid G_t) + \beta\, s_\phi(a\mid G_t),
\label{eq:posterior_logits}
\end{equation}
where $a \in \mathcal{A}_t^K$ and $\beta \ge 0$ controls the prior strength. The parameter $\beta$ controls the influence of the risk prior on the final action distribution, with larger values placing more weight on prior-favored actions relative to the learned policy logits. 

Then we have the candidate-restricted policy which is defined as:
\begin{equation}
\pi_{\theta,\phi}(a\mid G_t,\mathcal{A}_t^K)
=
\mathrm{softmax}_{a\in\mathcal{A}_t^K}\big(\tilde{\ell}(a\mid G_t)\big).
\label{eq:posterior_policy}
\end{equation}
Thus, the candidate set is determined by the risk prior, and the final action probabilities are obtained by reweighting the learned logits with the same risk score.

\subsection{Training}
\label{sec:training}
We train the method in two stages. We first collect a dataset of hazard observations and one-step action outcomes. For a hazard state $o_t$ and an action $a$, we obtain the supervision target by one-step simulation, and then train the risk surrogate with the $L_2$ loss. After this stage, the risk score network $s_\phi(\cdot\mid G)$ is frozen.

We then train the policy value network with PPO (Proximal Policy Optimization \cite{schulman2017proximal}) on macro-steps indexed by $m$. At hazard time $t_m$, we construct $\mathcal{A}_{t_m}^K$ via \eqref{eq:topk}, and then sample as:
\begin{equation}
a_m \sim \pi_{\theta,\phi}(\cdot\mid G_{t_m},\mathcal{A}_{t_m}^K).
\end{equation}
We then execute that action once and continue advancing the environment under the default non-hazard policy until the next hazard time, episode termination, or rollout segment truncation. This yields the macro-reward $R_m$ in \eqref{eq:macro_reward} and the macro-discount $\Gamma_m=\gamma^{k_m}$.
Over a collected rollout segment, we compute semi-MDP returns recursively:
\begin{equation}
\hat{G}_m
=
\begin{cases}
R_m, & \text{if the macro-transition terminates},\\[4pt]
R_m + \Gamma_m \hat{G}_{m+1}, & \text{otherwise.}
\end{cases}
\end{equation}
If the final macro-transition in the rollout segment is truncated before termination, we bootstrap its target with the value estimate of the truncation state as: $\hat{G}_m = R_m + \Gamma_m V_\theta(G_{\mathrm{trunc}})$. We then form the advantage estimates as: $\hat{A}_m = \hat{G}_m - V_\theta(G_{t_m})$, and normalize them over the rollout batch.
PPO is applied to the candidate-restricted policy with the standard clipped surrogate objective, which is defined as: 
\begin{equation}
\mathcal{L}_{\mathrm{PPO}}(\theta)
=
\mathbb{E}\left[
\min\!\left(
r_m(\theta)\hat{A}_m,\;
\mathrm{clip}\!\left(r_m(\theta),1-\epsilon,1+\epsilon\right)\hat{A}_m
\right)
\right],
\end{equation}
where
\begin{equation}
r_m(\theta)
=
\frac{
\pi_{\theta,\phi}(a_m\mid G_{t_m},\mathcal{A}_{t_m}^K)
}{
\pi_{\theta_{\mathrm{old}},\phi}(a_m\mid G_{t_m},\mathcal{A}_{t_m}^K)
}.
\end{equation}
The full training loss augments the clipped PPO term with value regression towards $\hat{G}_m$ and entropy regularization, as in \cite{schulman2017proximal}.
\section{EXPERIMENTAL RESULTS}
\label{sec:experiments}

We evaluate the proposed method on the environments  introduced in the L2RPN obtained from the Grid2Op framework \cite{grid2op}. We analyze each of them in the following section. For each environment, we report \textit{(i)} the average episodic reward, \textit{(ii)} the average number of steps survived, and \textit{(iii)} the average decision time in milliseconds per step. The latter captures the full online inference cost, including candidate construction and any optional screening stage, and is therefore the most relevant metric for assessing practical deployment under real-time constraints. We compare our method against \textit{(i)} a Random policy, which uniformly samples an action from the action set, and \textit{(ii)} a PPO baseline, which uses the learned policy without the proposed risk-informed prior. We also report a Greedy oracle that selects actions through exhaustive one-step simulation over candidate controls. For \texttt{case118}, we replace the Greedy oracle with the publicly available LJN agent as a baseline \cite{pavao2025ai}, an adaptation of the winning solution to the L2RPN 2023 IDF AI challenge that combines and extends earlier strong approaches with safety oriented heuristics to operate the grid in a cost effective way, while maximizing renewable energy use. In our setting, we use its topology-only variant. This framework is available only for that environment and represents a particularly strong, highly specialized engineering solution tailored to the specific action space. However, LJN cannot be reused as is for other environments and would require regenerating the action space entirely which comes with a huge computational cost.

\subsection{Datasets}

We evaluate our approach on three benchmark environments of increasing scale and operational complexity. 
The first environment, \texttt{case14}, is derived from a slightly modified IEEE 14-bus system and includes 14 substations, 20 transmission lines, 6 generators, and 11 loads. This environment does not include storage units and is commonly used as a compact testbed for topology-control algorithms. 
The second environment, \texttt{case36}, is a 36-substation benchmark extracted and modified from an IEEE 118-bus system. In contrast to \texttt{case14}, it enables maintenance constraints (scheduled outages and associated operational restrictions) while excluding both storage and adversarial perturbations, thereby emphasizing constrained operation under limited topological flexibility. 
The largest environment, \texttt{case118}, is again based on a modified IEEE 118-bus system, which is intended to approximate a prospective French power-system configuration around 2035. In particular, the generation portfolio is adjusted to increase the contribution of variable renewable energy sources (e.g., wind and solar), which increases the variability of net injections and makes balancing and congestion management more challenging. The \texttt{case118} setting further includes \textit{(i)} maintenance constraints, \textit{(ii)} storage devices, and \textit{(iii)} an adversarial opponent that models malicious or extreme exogenous events. Operationally, the opponent can trigger forced disconnections (attacks) on up to three transmission lines per time step, subject to geographic partitioning of the network into three areas and a constraint of at most one attacked line per area, thereby inducing non-local flow redistributions and potentially cascading overloads. Correspondingly, the action rules permit corrective interventions on up to three lines and three substations per step, again limited to one action per area, to reflect realistic operational constraints on concurrent switching and to prevent unrealistically large instantaneous topology changes. For each case, we split the available episodes into disjoint training and evaluation sets. For evaluation, we reserved five unseen episodes of 2017 micro-steps with 5 minutes resolution, which means 7 days of continuous operation.

\subsection{Experiments}

\begin{table}[t]
\vspace*{8pt}
\centering
\caption{Results on \texttt{case14}
($|\mathcal{A}|=209$, $|\mathcal{A}_t^K|=64$).}
\label{tab:case14_train}
\small
\setlength{\tabcolsep}{3pt}
\begin{tabularx}{\columnwidth}
  {@{}>{\raggedright\arraybackslash}Xrrr@{}}
\toprule
Method
& \shortstack[r]{Avg.\\reward}
& \shortstack[r]{Avg.\\steps}
& \shortstack[r]{Decision time\\(ms/step)} \\
\midrule
Random        & 28.9     & 1.6    & 0.008 \\
PPO           & 106856.1 & 1679.0 & 0.009 \\
Greedy oracle & 127822.8 & 2017.0 & 0.190 \\
\textbf{Ours} & 127822.5 & 2017.0 & 0.032 \\
\bottomrule
\end{tabularx}
\end{table}

\begin{table}[t]
\centering
\caption{Results on \texttt{case36}
($|\mathcal{A}|=66{,}978$, $|\mathcal{A}_t^K|=512$).}
\label{tab:wcci2020}
\small
\setlength{\tabcolsep}{3pt}
\begin{tabularx}{\columnwidth}
  {@{}>{\raggedright\arraybackslash}Xrrr@{}}
\toprule
Method
& \shortstack[r]{Avg.\\reward}
& \shortstack[r]{Avg.\\steps}
& \shortstack[r]{Decision time\\(ms/step)} \\
\midrule
Random        & 1138.1   & 42.6   & 0.005   \\
PPO           & 120540.6 & 1390.8 & 4.081   \\
Greedy oracle & 158549.3 & 1866.8 & 184.199 \\
\textbf{Ours} & 149963.1 & 1774.2 & 0.924   \\
\bottomrule
\end{tabularx}
\end{table}

\begin{table}[t]
\vspace*{8pt}
\centering
\caption{Results on \texttt{case118}
($|\mathcal{A}|=73{,}357$, $|\mathcal{A}_t^K|=512$).}
\label{tab:case118}
\small
\setlength{\tabcolsep}{3pt}
\begin{tabularx}{\columnwidth}
  {@{}>{\raggedright\arraybackslash}Xrrr@{}}
\toprule
Method
& \shortstack[r]{Avg.\\reward}
& \shortstack[r]{Avg.\\steps}
& \shortstack[r]{Decision time\\(ms/step)} \\
\midrule
Random              & 1038.1   & 13.2  & 0.006  \\
PPO                 & 23830.9  & 113.4 & 18.477 \\
LJN (topology-only)  & 137420.6 & 717.6 & 81.495 \\
\textbf{Ours}       & 84643.5  & 435.4 & 32.405 \\
\bottomrule
\end{tabularx}
\end{table}

Tables~\ref{tab:case14_train}--\ref{tab:case118} summarize the performance on \texttt{case14}, \texttt{case36} and on \texttt{case118} in terms of average reward, average number of steps, and decision time per step. Since the strongest reference method is not the same across all benchmarks (LJN on \texttt{case118}, and the Greedy oracle on \texttt{case14} and \texttt{case36}), and the environments present different challenges, the most meaningful comparisons are within each case rather than across cases. Overall, the proposed method consistently outperforms both Random and PPO in reward and episode length on all three systems, indicating that the additional decision mechanism improves control quality beyond a standalone learned policy. The results also show a clear relationship between reward and survivability, since the methods that keep the system operational for more steps consistently obtain larger cumulative rewards.

On \texttt{case14}, the proposed method is essentially indistinguishable from the Greedy oracle. It achieves an average reward of 127822.5, compared with 127822.8 for the oracle, and matches the oracle exactly in average episode length (2017.0 steps). This is a strong result, as it suggests that on the smallest benchmark the proposed approach can recover near-oracle behavior without relying on expensive search. Although the decision time of our method (0.032 ms/step) is higher than that of PPO (0.009 ms/step), the absolute overhead remains negligible, and it is still about 6$\times$ faster than the Greedy oracle. In other words, the proposed method achieves near-oracle performance while remaining highly efficient.

On \texttt{case36}, the task becomes more challenging, but the same overall trend remains. The proposed method reaches an average reward of 149963.1 and 1774.2 average steps, corresponding to gains of about 24.4\% in reward and 27.6\% in episode length over PPO. The Greedy oracle still achieves the best absolute performance on this case, but the gap is moderate: our method reaches about 94.6\% of the oracle reward while requiring far less computation. In particular, the Greedy oracle needs 184.199 ms/step, whereas the proposed method requires only 0.924 ms/step, i.e., roughly 200$\times$ less decision time. Interestingly, our method is also faster than PPO on this benchmark, so the performance improvement does not come at the cost of slower inference.

On \texttt{case118}, the proposed method again substantially improves over PPO, increasing the average reward from 23830.9 to 84643.5 and the average number of steps from 113.4 to 435.4. This corresponds to improvements of about 255\% in reward and 284\% in survived steps, showing that the additional decision component becomes particularly beneficial on the larger system. At the same time, the specialized LJN topology-only baseline remains stronger on this benchmark, with 137420.6 reward and 717.6 steps. Even so, the proposed method reduces decision time from 81.495 to 32.405 ms/step, making it about 2.5$\times$ faster than LJN. These results suggest that, on the largest case, there is still room to improve the quality of topology decisions, but the proposed method already offers a much better trade-off than PPO and a substantially cheaper alternative to the stronger topology-based baseline. Notably, these results are obtained while restricting each hazardous intervention to a single unitary topology action; this conservative action budget keeps online decision costs low, but it  also explains part of the remaining gap to the best performer, where the benchmark permits multiple concurrent interventions per step.

Overall, these results show that our proposed method provides a strong balance between control quality and computational cost across the three benchmarks. Essentially, it matches the oracle on \texttt{case14}, remains close to the oracle while being dramatically faster on \texttt{case36}, and delivers much larger gains over PPO on \texttt{case118} while reducing the cost relative to the topology-based baseline. By contrast, the Random baseline performs poorly in every setting, confirming that this problem requires structured and informed decision making. Overall, the experiments support the conclusion that combining policy-based control with our proposed decision mechanism improves both robustness and long-horizon performance while keeping inference fast enough for practical deployment.
\section{CONCLUSION}

We introduced a risk-informed reinforcement learning approach for topology control in power grids that addresses two central challenges of the problem: the combinatorial size of the action space and the high computational cost of online action screening. The proposed framework combines a semi-Markov control with a learned Gibbs prior over actions, where a graph neural network surrogate predicts post-action overload risk and is used both to construct a small state-dependent candidate set and to reweight the policy scores. This yields a control architecture that focuses computation on safety critical states and biases action selection towards physically plausible interventions.

Experiments on three realistic environments show that our method provides a strong balance between control quality and runtime. It essentially matches the Greedy oracle on \texttt{case14}, remains close to the oracle while being dramatically faster on \texttt{case36}, and substantially improves over PPO on the more challenging \texttt{case118} environment while also reducing decision time relative to the specialized LJN baseline. 

A direction for future work is to improve the quality of the risk surrogate on larger environments such as \texttt{case118}, where stronger topology specialized methods still achieve better absolute performance. A further research direction is to extend the framework to continuous action spaces by learning a direct policy for control variables, thereby avoiding the need for iterative numerical optimization to compute actions such as optimal dispatch or curtailment at each decision step.

\bibliographystyle{IEEEtran}
\bibliography{refs}

\end{document}